# Feature Selection on Noisy Twitter Short Text Messages for Language Identification

**Mohd Zeeshan Ansari\*, T Ahmad, A Fatima**
**Department of Computer Engineering, Jamia Millia Islamia, New Delhi.**

## Abstract
The task of written language identification involves typically the detection of the languages present in a sample of text. Moreover, a sequence of text may not belong to a single inherent language but also may be mixture of text written in multiple languages. This kind of text is generated in large volumes from social media platforms due to its flexible and user friendly environment. Such text contains very large number of features which are essential for development of statistical, probabilistic as well as other kinds of language models. The large number of features have rich as well as irrelevant and redundant features which have diverse effect over the performance of the learning model. Therefore, feature selection methods are significant in choosing feature that are most relevant for an efficient model. In this article, we basically consider the Hindi-English language identification task as Hindi and English are often two most widely spoken languages of India. We apply different feature selection algorithms across various learning algorithms in order to analyze the effect of the algorithm as well as the number of features on the performance of the task. The methodology focuses on the word level language identification using a novel dataset of 6903 tweets extracted from Twitter. Various n-gram profiles are examined with different feature selection algorithms over many classifiers. Finally, an exhaustive comparative analysis is put forward with respect to the overall experiments conducted for the task.

## 1. Introduction
Language identification (LI) is the significant area of research from several decades and has become a challenging problem after the dramatic increase in the data generated by social media where a large mass of socially connected people use an informal languages for communication. Language identification problem is well studied as canonical text formulation in spoken as well as textual forms, however, identification of languages of documents in multilingual text is still facing challenges. LI is the preprocessing step for certain natural language processing tasks like machine translation, information retrieval, information extraction, sentiment analysis, etc. In work, we study this problem specifically as an exploratory analysis of social media content on the basis of language. The social media platforms like Facebook, Twitter and WhatsApp are quiet popular for information sharing and the countries like India is a place for multilingual community with diverse languages. India has 22 official languages along with several minority languages are present across the country.

A feature selection methods largely affects the quality of the input features, which in-turn affects the performance of classification methods. The feature selection is generally illustrated as a subset selection problem in context of classification, amongst which the common feature selection algorithms are forward feature selection and backward elimination. The other general feature selection approach may be characterized in two significant steps (1) prepare a ranked list of features in the dataset using a scoring functions and (2) select the top top-k features from the ranked list, where k may be the significant number of features essential for classification. For our experiments, twitter is considered as a data source and we specifically address the problem of language identification of code mixed social media text. The feature selection algorithms which are applied on the language identification tasks are forward feature selection, backward elimination and top-k ranked. The experiments with different feature selection algorithms across various classification methods are carried out and the analysis of results is presented.

## 2. Literature Survey

Language Identification task was established as a classification problem until it was popularly recognized by Beesley (1988) and subsequently as a text categorization problem through textcat tool of Canvar and Trenkle (1994) based on character frequency model [1,2]. It was thoroughly studied as code mixing task by Das and Gamback (2014) as they proposed the code mixing index for the mixing of languages in facebook posts and study the problem by using n-grams and weighted n-grams as features along with support vector machines [3]. Several statistical models applied to Language Identification include Bayesian models (Dunning 1994, Grefenstette, 1995, Lui and Baldwin, 2011), kernel methods and support vector machines by Teytaud and Jalam (2001) and Lodhi et al. (2002) [ 4-8]. Weakly supervised models used by King and Abney (2013) perform word level language identification [9]. Code-mixing and code-switching is predominantly observed in large number of language pairs, the users native language is primarily the base language with inclusions of English words. Deepthi et al. (2018) et al. studied code switching patterns for Hindi-English and Spanish-English language pairs and showed that CRF model performed better than neural networks by 3-5 percent on the twitter dataset [10]. Gupta et.al. (2018) studied the effect of n-grams profiles in identification of Indo-Aryan languages using word-level LSTM, subsequently, obtained the F1 score of 0.836 and concluded that performance increases on increasing the size of n-gram [11]. Ranjan (2016) used deep back propagation network for the language identification on limited size of training data along with stochastic gradient descent with a batch size of 256 [12]. Maharjan (2015) investigated the Spanish-English and Nepali-English language pairs on Twitter data using Conditional Random Fields with Generalized Expectation comparing with dictionary based approaches [13]. The CRF outperforms the dictionary based approaches with an accuracy of 86%. Sequiera (2015) studied the problem of language identification over three different subtasks. They performed the labeling of words of code switched text for 8 different Indian languages. They also tried to retrieve the documents including astrology, movie reviews, lyrics etc. written over both Devanagri and Roman transliterated Hindi documents [14]. Kriz et al. (2015) studied the identification of English language by the native writers when used by non-native writers. The classification method with cross entropy measure of four features tokens, characters, suffixes and POS tags achieved a 10 fold cross validation an accuracy of 82.4 % [15].

The Discriminating Similar Languages (DSL 2015) shared task focused on building systems for distinguishing language pairs. The task used the training data of 20000 sentences from 13 different languages and 1000 unlabeled sentences for evaluation from each language. Mathur et al. (2015) applied Multinimial Naïve Bayes, Logictic Regression and ensemble methods and obtained accuracy of 90% between many similar languages [16]. Bali et al. (2014) performed the analysis of code-mixing by collecting the facebook pages of Bollywood celebrities and BBC and concluded that the use a greater mix of Indian languages with Hindi [17]. Barman et al. (2014) performed the language identifcation on Bengali, Hindi and English code-switched data in which he employed unsupervised dictionary based approach, supervised word level classification and a sequence labeling approach using CRF [18]. Solorio et al. (2014) studied language identification for four language pairs-Spanish-English, Mandarin-English, Modern Standard Arabic Dialectal Arabic, Nepali-English. Results showed that language identification of code switched closely related languages is more difficult at the token level and needs further attention [19]. Jiang et al. (2014) studied the problem of spoken language identification utilizing deep bottleneck features addresses the problem of labeling individual words in multilingual document using weakly supervised methods. In their work conditional random field performed consistently well [20]. Earlier notable works include the work by Baldwin and Lui (2010) where they used three different datasets for language classification [21]. They conducted experiments using nearest-neighbor and nearest prototype model, Naïve Bayes and Support

Vector Machines. They used byte and codepoints as tokenization strategies and conducted experiments n grams. Lui and Baldwin (2011) used the cross domain feature selection approach on language identification task. They used byte n-grams as features with information gain as the selection criteria. The features having high information gain in context to language and low information gain in context to domain are selected [6].

## 3. Feature Selection Methods

Character N-Grams are efficient and language independent useful for several text classification problems [2]. These are quiet relevant in capturing morphological characteristics of a word especially prefixes, suffixes and subwords such as *ing* in case English words, eg *playing*, *going* and *na* in case of Hindi words eg. *khelna*, *chalna*. Also, they have proven helpful in capturing semantics in noisy data of code mixed language [6, 18] Character N-Grams for n = 1 to 4 are considered for the selection of feature features. Feature or variable selection is necessary for enabling data understanding and reducing computational requirements by predominantly defying the curse of dimensionality [22]. A typical learning task may involve several thousands of features as in named entity classification of Wikipedia articles in which automatically choosing the relevant features may dramatically improve the prediction performance [23,24]. The significant repercussion of learning a powerful model with large number of irrelevant features is reduced model performance [25].

### 3.1 Top-k selection

Selecting top-k features requires a scoring function as a parameter which it applies on the given dataset by taking pairwise each feature with the output. The scoring function returns scores for each pair of input-output pair, the first k features are selected according to the scored highest among all feature scores. For our work we have used chi-square test as the parameter for scoring function.

### 3.2 Forward feature selection

This algorithm calculates the accuracy of model using every feature sequentially. A single features with best accuracy are chosen and kept in the selection set. Now every other feature is combined with previously selected feature in the set in order to select the second best subset. This process continues until the specified number of features are obtained in the selection set. This process requires excessive computational cost and hence certain optimization technique may also be applied to reduce the computations. Sequential feature selector contains a number of parameters. Sequential feature selection can be forward or backward. For some researchers forward feature selection is more efficient in terms of computation than backward feature selection while generating the feature subset. But, some researchers argue that in case of forward selection the significance of the feature selected is not taken in context of other variables that are yet to be added. This can be better explained via considering an example. Suppose, we have three variables. The forward feature selection picks up a single variable that makes the best classification at that step. If we take two variables together for classification, it has been found that the pair of those variables that were discarded at the initial step performs the best classification than the best selected feature taken with any of those two variables. However, in case of backward selection the first step is to eliminate that best performing variable to retain the pair of other two variables that are providing best results. But if we have to consider the single variable for this task, the backward elimination will end up in loosing that feature in the initial task.

### 3.3 Recursive feature elimination

The recursive elimination operates in backward direction i.e. it initially considers the entire feature set and selects the specified number of features by recursively eliminating weakest features. However, in doing so it uses the feature ranking methods which are generally done using the feature importance according to the classifier coefficients. In the first step, the model is fitted on linear regression model considering all the available features. Next, it calculates the coefficients for linear regression model, which act as the ranking criteria for those features, subsequently, removing the variable with least ranking value. This process is recursively repeated until the required number of features are obtained. Sometimes the feature count to select is not explicitly specified, in that case, the cross validation is applied with recursive elimination to rank different feature subsets and then choosing the subset with best cross validation score.

## 4. Classification Methods for Language Identification

Classification is the methodology to segregate the data into a desired and distinct number of classes where each class is assigned a unique class label. In the vocabulary of machine learning, classification comes under the category of supervised learning, i.e., learning using correctly labeled data points. Models or algorithms that perform the job of classifying data instances into different classes and provide them a discrete class label are termed as classifiers. They are generally approximation functions that operate on the given input points to produce the label class. The input set consists of certain attributes or features that describe the properties for classes. The individual observations usually contains values that are categorical (e.g. "A", "B", "AB" or "O", for blood type), ordinal (e.g. "high", "low"), integer-valued (e.g. word count for documents) or real-valued (e.g. temperature). The observations are termed as instances, the attributes are known as features which are grouped into a feature vector, and the labels to be assigned are the classes. The different classification algorithms used in this work are Logistic Regression, Multinomial Naïve Bayes and Decision Tree.

### 4.1 Logistic Regression

At its core, logistic regression is based on the logistic function, which is also termed as sigmoid function. This curve is known as the sigmoid curve, is the function that maps any values between 0 and 1. Different variations of logistic regression are binomial and multinomial logistic regression. For binomial logistic regression the dependent variable can have two possible outcome i.e. '0' or '1'. In case of multinomial logistic regression the dependent variable can take three or more values as in our work the language labels are classified as 'Hindi', 'English' and 'Other'.

### 4.2 Multinomial Naïve Bayes

Naive Bayes classifier which is based on Bayes Theorem is popularly used for classification in which it is assumed that all the features within a feature set are conditionally independent .Naive Bayes is a probability based classifier where the probability of a data point belonging to a particular class is defined in terms of the likelihood of feature and class prior probability. It is one of the most popular model used for document classification task.

### 4.3 Decision Tree

Decision tree classifiers fall under the category of supervised classification algorithms, where a decision tree has to be taken at each node to perform the split. Nodes can be considered as a feature and splitting criteria decides the threshold for splitting into different classes.

## 5. Methodology

The dataset preparation from Twitter is carried out for this work in order to evaluate the Language Identification task as Twitter is the rich source of code mixed text. There are several sources for the preparation of data, perhaps, Other than social media content, Wikipedia is a significant source of textual data and used in several applications. The text collected from Twitter is cleaned and preprocessed and

| Table 1 | |
|---|---|
| Type of tweets | Count |
| Hindi English code mixed | 3854 |
| English | 3049 |
| Total | 6903 |

annotated at word level with three labels English, Hindi and Named Entity. The labelled data is classified on Logistic Regression, Multinomial Naïve Bayes and Decision Tree classifiers along feature selection using Top-k, Forward Selection and Backward elimination techniques.

**5.1 Corpus Acquisition**
The corpus acquisition is based on the twitter handles of those persons and organizations that often tweet in Hindi written using Roman script due to their richness in code mixing of languages. While manually examining the twitter text space, it was observed that code mixing is occurring very frequently in the tweets generated by some specific twitter handles. Such twitter handles were carefully identified and chosen for this task, hence specifically contain the Hindi English code mixed tweets. Mainly, the selected handles are @hinglishnews, @TheHinglish, @Sonytv, @Zeetv, @KapilSharmaK9, @BeingSalmanKhan and @DrKumarVishwas. These code mixed tweets consists majority of transliterated Hindi words with very small number of pure English. In order to balance the distribution of both tweets of English and Hindi words in corpus, pure English tweets were also collected from the twitter handles of @the hindu @ShashiTharoor and @BarackObama. Finally, a corpus of 6903 tweets containing a mix of transliterated Hindi and English words is obtained.

**5.2 Preprocessing**
Before the extraction of feature, the corpus is thoroughly pre-processed according to (1) removal of the web links, URLs, hashtags, emoticons, punctuations, twitter handles (2) removal of words in language other than English (3) replacement of hashtags with its normal text (4) case folding (5) removal of non-informative words using English stop words list (6) removal of redundant words from 16745 words

**5.3 Annotation**
Single layer annotation is carried out by assigning the word level tags according to the tagset displayed in Table X leading to 13294 unique words.

| Table 2 | | | | |
|---|---|---|---|---|
| Tag | Description | Count | %age | Example word |
| EN | English word | 8298 | 49.6 | *dhamaal, bharat, ko, mila, dusra* |
| HI | Transliterated Hindi word | 3042 | 18.2 | *silver, pilot, tax* |
| NE | Named Entity | 1954 | 11.7 | *Mohindar. Allahabad* |
| - | Redundant words | 3451 | 20.5 | - |
| - | Total | 16745 | 100 | - |

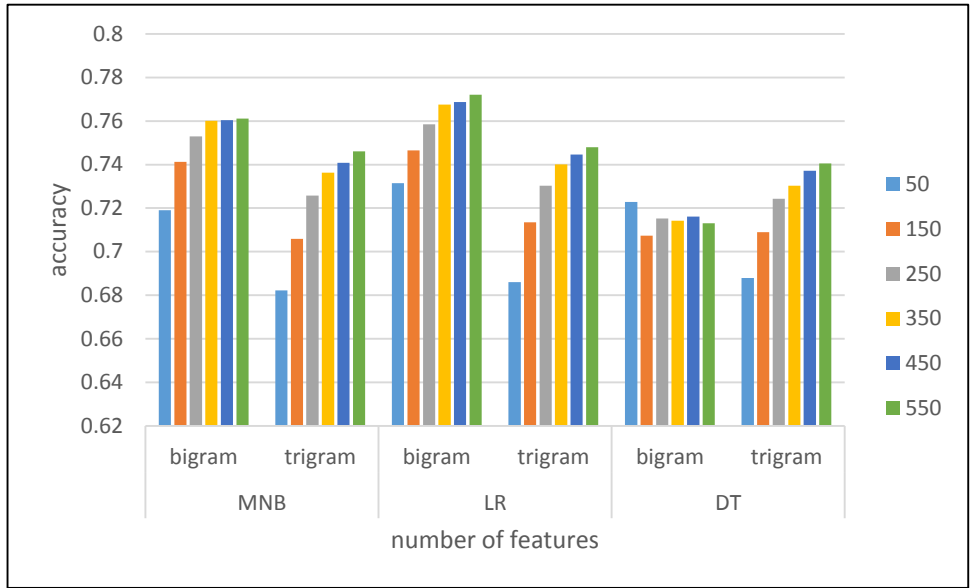

*Figure 1 Performance of selected bigram and trigrams features using top-k feature selection across different classifiers*

**5.4 Experimental setup**

The section describes the analysis of results obtained after the application of aforesaid methodology. All analysis are based on the experiments conducted on the code mixed dataset prepared from twitter. Figure 1 shows the accuracy of the Forward Feature Selection method on bigram and trigram profiles using Naïve Bayes, Logistic Regression and Decision Tree classifier versus the number of features selected. The x-axis denotes the number of features selected based on the type of feature selected against each classifier and the y-axis denotes the accuracy of the classifier.

# 6. Analysis of Results

Fig. 1 shows the performance of top-k feature selection method over bigram and trigram features for number of features selected 50 through 550 across three classifiers. It is clearly observed from this table that on increasing the number of features selected, the accuracy increases which is in line with the principle of feature selection. However, it also shows exceptional behavior in case of bigrams on decision tree.

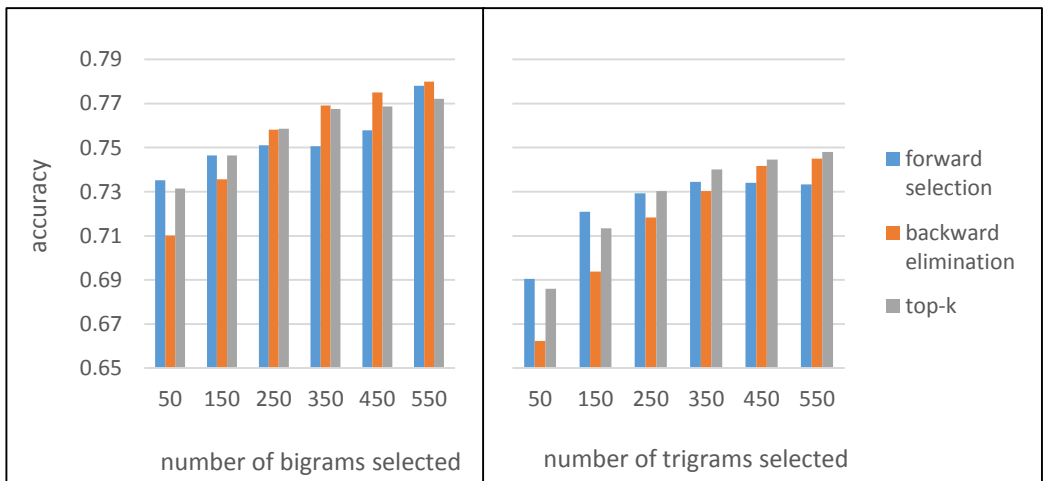

*Figure 2 Accuracy in respect of bigrams and trigrams for forward selection, backward elimination and top-k selection.*

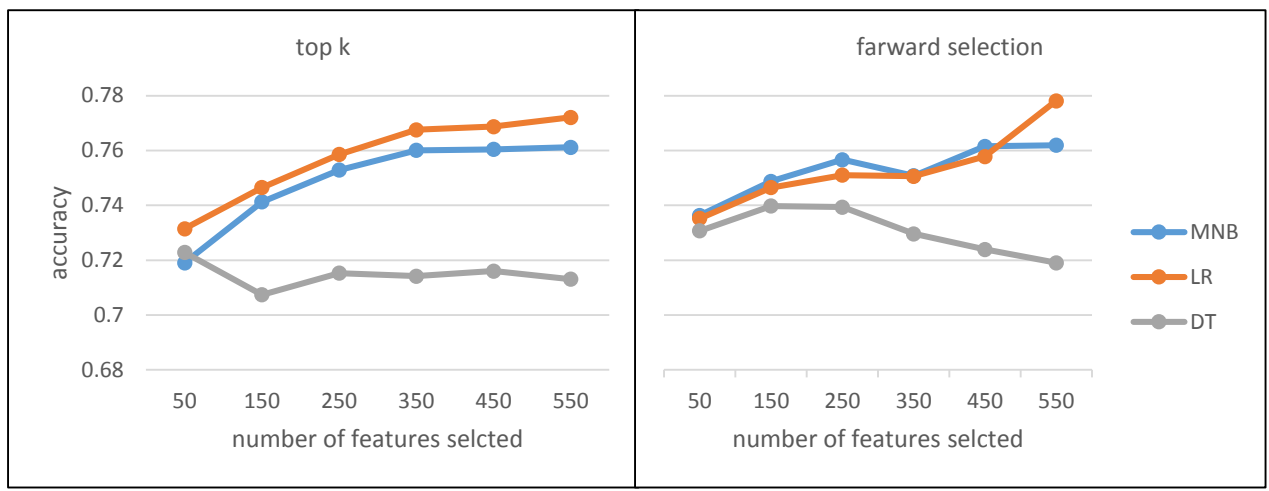

*Figure 3 Accuracy curves of top-k and farward feature selection when classified using Logistic Regression*

Secondly, the observation that can be derived from Fig. 1 is that there is higher rate of increase in accuracy from 50 to 250 selected features however, this rate slows as number of features selected in increased from 350 till 550. Thirdly, it can be concluded that the best classification accuracy of top-k feature selection method is obtained for all bigram features selected is 0.77 when classified with Logistic Regression classifier. Each bar in the Fig. 2 denotes one of the following used feature selection algorithms with different number of selected features: (1) Forward Selection (2) Backward elimination and (3) Top-k. The comparative performance of feature selection algorithms with forward selection and backward elimination is shown in Fig 2(a) with respect to bigram features and in Fig 2(b) with respect to trigram features respectively. Fig 3 shows the performance the curves of accuracies obtained in respect of different classifiers when features are selected using top-k and forward selection. The Logistic Regression classifier shows the best performance in top-k selection and partially in case of forward feature selection.

## 7. Conclusion

Feature selection approaches are useful over language classification problem if the number of features are large. The trade-off between accuracy and selected number of features can be pruned by experiments. We have performed feature selection over three methods. We can conclude from our experiments that logistic regression with forward selection outperforms top-k selection which in turn outperforms all the other models for the task. For future work, the evaluation of the combination of filter and wrapper methods for language classification can be considered for improvement. Filter method may be applied as preprocessing step to remove features with no relation to the language classification model and finally the wrapper method may collect the best set of features with respect to the learning model. Finally, Language Identification may also be proven appropriate task for other feature selection approaches with several classification methods.